\documentclass[letterpaper, 10 pt, conference]{ieeeconf}  % Comment this line out if you need a4paper

\IEEEoverridecommandlockouts                              % This command is only needed if 
\usepackage{amsfonts}
\usepackage{amsmath}
\usepackage{bm}
\usepackage{amsmath,amssymb}
\usepackage{mathtools}
\usepackage{color}

\title{\LARGE \bf
Cell Tracking in {\it \bf C. elegans}\\
with Cell Position Heatmap-Based Alignment and Pairwise Detection
}

\author{Kaito Shiku$^{1}$, Hiromitsu Shirai$^{1}$, Takeshi Ishihara$^{1}$, and Ryoma Bise$^{1}$% <-this % stops a space
\thanks{$^{1}$ with Kyushu University, Fukuoka, Japan
        {\tt\small kaito.shiku@human.ait.kyushu-u.ac.jp}}%
}

\begin{document}

\maketitle
\thispagestyle{empty}
\pagestyle{empty}

%%%%%%%%%%%%%%%%%%%%%%%%%%%%%%%%%%%%%%%%%%%%%%%%%%%%%%%%%%%%%%%%%%%%
\begin{abstract}
3D cell tracking in a living organism has a crucial role in live cell image analysis.
Cell tracking in {\it C. elegans} has two difficulties. First, cell migration in a consecutive frame is large since they move their head during scanning. Second, cell detection is often inconsistent in consecutive frames due to touching cells and low-contrast images, and these inconsistent detections affect the tracking performance worse.
In this paper, we propose a cell tracking method to address these issues, which has two main contributions.
First, we introduce cell position heatmap-based non-rigid alignment with test-time fine-tuning, which can warp the detected points to near the positions at the next frame. Second, we propose a pairwise detection method, which uses the information of detection results at the previous frame for detecting cells at the current frame. The experimental results demonstrate the effectiveness of each module, and the proposed method achieved the best performance in comparison.
%\newline
%\indent \textit{Clinical relevance}— This is a brief additional statement on why a this might be of interest to practicing clinicians. Example: This establishes the anesthetic efficacy of 10\% intraosseous injections with epinephrine to positively influence cardiovascular function.
\end{abstract}

%%%%%%%%%%%%%%%%%%%%%%%%%%%%%%%%%%%%%%%%%%%%%%%%%%%%%%%%%%%%%%%%%%%%%%%%%%%%%%%%
\section{INTRODUCTION}
3D cell tracking in a living organism is a fundamental task of live cell image analysis.
For example, tracking neuron cells in {\it C. elegans} is important to analyze their nervous activity.
In the study of neurons, a {\it C. elegan} is stimulated, and the neuron activities are captured by 3D microscopies, such as confocal microscopes. To analyze the temporal activity of neurons, cell tracking is required.

\begin{figure}[t]
 \begin{center} \includegraphics[width=0.9\columnwidth]{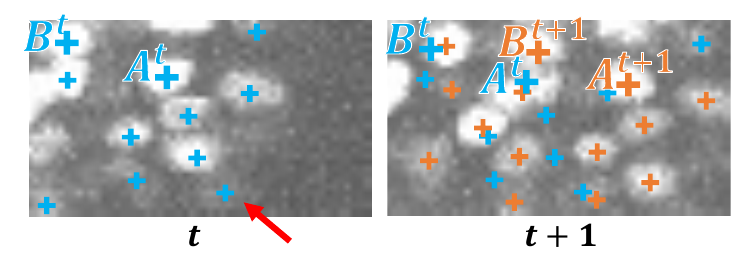}
    \caption{Example of large movement issue. Left: detection results at $t$ (blue `+'). Right: those at $t+1$ (orange `+').}
    \label{fig:intro}
 \end{center}
 \vspace{-5mm}
\end{figure}

Cell tracking in {\it C. elegans} has two difficulties.
First, cells often move large distances since they move their head due to stimulation, compared to 2D cell tracking in {\it vitro}.
As shown in Fig. \ref{fig:intro}, the distance of corresponding cells between consecutive frames is often larger than that between non-corresponding cells, e.g., $A^{t}$ is closer to $B^{t+1}$ than $A^{t+1}$. This makes it difficult to associate cells between frames based on their proximity.
Second, cells in microscopy often have low contrast (e.g., the cell indicated by the red arrow in Fig. \ref{fig:intro}). In such images, cell detection from a single input image often produces inconsistent detection in consecutive frames since it is difficult to identify such low-contrast cells only from a single frame. Inconsistent cell detection in consecutive frames may cause tracking errors.

This paper proposes a cell tracking method to address these difficulties, which has two main technical contributions. First, to address the large displacement of cells, we introduce cell position heatmap-based alignment with test-time fine-tuning.
This can estimate the displacement maps and warp cell positions from the current frame to the next frame. Using these warped positions, cell association can work appropriately while large displacement.
Second, to produce consistent detection in consecutive frames, we propose pairwise detection, which jointly inputs the original image at $t+1$ and the warped heatmap from $t$ to $t+1$, in which the warped map indicates the estimated current positions. It is expected that the detection results at $t+1$ contain the corresponding cells that were detected at the previous frame $t$ by the pairwise input.
In the experiments, we evaluated the tracking performance using real biological images. The experimental results demonstrate the effectiveness of each module, and the proposed method achieved the best performance in comparison.

Our main contributions are summarized as follows:
\begin{itemize}
      \item We introduce deep alignment of the cell position heat maps, which estimate the movement of cells in consecutive frames. The tracking accuracy was improved significantly using this displacement information. 
      \item To obtain accurate alignment, we used cell position heat maps instead of original images, and we introduce test-time fine-tuning.
      \item We propose a pairwise detection to produce consistent detection in consecutive frames.
\end{itemize} 

\section{Related works}

\subsection{2D cell Tracking}
Many cell tracking methods in 2D microscopy images take a tracking-by-detection approach, in which individual cells are detected in each frame, and then the detected cells are associated in consecutive frames using kalman filter \cite{huth2010significantly}, linear programming \cite{kanadeT2011,bise2013,zhou2019joint,WuZ12} and graph-based optimization \cite{bise2011reliable,Funke18}. In such methods, the detection is separate from the tracking.
Hayashida et al. \cite{hayashidaJ19,hayashidaJ20} proposed tracking methods for jointly estimating the position and motion between two frames. These methods are more robust for large displacement, however, these may overfit to training data, and they cannot be fine-tuned for test data since it requires the ground truth of cell motion information for training. Unlike these methods, our method can re-train the network without supervised data.

\vspace{-20mm}

\subsection{3D Cell Tracking in {\it C. elegans}}
Many cell tracking methods in {\it C. elegans} also take a tracking-by-detection approach.
For example, Lauziere et al. \cite{lauziere2022semi} proposed a tracking method that detects cells by 3D U-net \cite{ronneberger2015u}, and associates detections with linear programming. However, this method does not work properly when cell movement is large. Chen et al. \cite{chen2015c} use relative position information among neighbor cells for cell association, which facilitates associating cells when large displacement. This method assumes that cells are detected successfully in each frame. If false negatives occur at a frame, it affects the relative position relationship, and it may cause tracking errors.
To address such difficulties, we introduce cell position heatmap-based alignment in consecutive frames, and we propose a pairwise detection method, which jointly inputs the original image at $t+1$ and the warped cell position heatmap at the previous frame.

\section{Deep non-rigid alignment-based cell tracking with consistent detection}
\begin{figure}[t]
 \begin{center} \includegraphics[width=0.8\columnwidth]{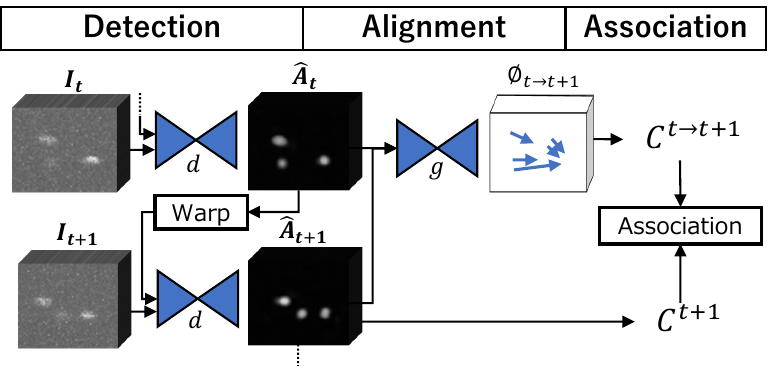}
    \caption{Overview of our method. First, given 
two consecutive frames of time-lapse images as inputs, a detection module estimates cell position at $t+1$ using cell position information at $t$.
Second, the alignment module warps cell positions at $t$ to $t+1$.
Finally, we associate the warped cell positions at $t$ and those at $t+1$.
}
    \label{fig:overview}
 \end{center}
  \vspace{-3mm}
\end{figure}

%\begin{figure}[t]
% \begin{center} \includegraphics[width=1.0\columnwidth]%{./figs/move.pdf}
%    \caption{$\Phi$}
%    \label{intro}
% \end{center}
%\end{figure}

As shown in Fig. \ref{fig:overview}, the proposed method consists of three steps: 1) cell position heatmap-based alignment, which aligns images in consecutive frames from time $t$ to $t+1$, and estimates the detected cell positions at the next frame by warping; 2) pairwise detection, which detects individual cell positions in a 3D volume using the current original image and the warped cell position heatmap; and 3) cell association, which associates the detected cells in consecutive frames using the warped positions. These steps are applied for each consecutive frame iteratively.

\subsection{Cell Position Heatmap-Based Alignment with Test-Time Fine-Tuning}
To address the large displacement of cells, we introduce deep alignment, which estimates the displacement maps and warps cell positions from the current frame to the next. These warped positions are used for both detection and association.

To perform this alignment, we take a deep non-rigid alignment with test-time fine-tuning using VoxelMorhph \cite{balakrishnan2019voxelmorph}, which trains a network to produce a displacement field so that the warped source image is similar to the target image, using the unsupervised sequential data.
For the alignment network, we use V-net \cite{milletari2016v}, which has a U-net \cite{ronneberger2015u} structure with 3D convolutional layers.

\subsubsection{Pre-training}
Given sets of consecutive images $I_t, I_{t+1} \in \mathbb{R}^{w\times h \times d}$ in a time-lapse sequence as training data, the network $g$ is trained to align the current image $I_t$ to the next time frame $I_{t+1}$. 
%this step aligns the current image $I_t$ to the next time frame $I_{t+1}$.

Let $\phi$ be a displacement field that transforms the coordinates of $I_t$ to those of $I_{t+1}$. 
The alignment network $g$ is trained by the entire loss defined as:
\begin{equation}
\label{eq:RegistrationLoss}
\mathcal{L} = \mathcal{L}_{sim}(I_{t+1},I_{t} \circ \bm{\phi}) + \gamma \mathcal{L}_{smooth}(\phi), 
\end{equation} 
where $I_t \circ \phi$ represents the aligned image of $I_t$ warped by $\phi$, function $\mathcal{L}_{sim}(\cdot,\cdot)$ measures image similarity between its two inputs, and $\mathcal{L}_{smooth}$ is a regularization term that enforces a spatially smooth deformation.
In our method, we used a MSE loss for $\mathcal{L}_{sim}$, and the spatial gradients of $\phi$ for $\mathcal{L}_{smooth}$, these are defined as follows: 
%defined as:
%and the spatial gradients of displacement $\phi$ for $\mathcal{L}_{smooth}$, these are defined as follows: 
\begin{equation}
\label{eq:SimLoss}
\mathcal{L}_{sim}(I_{t+1},I_{t} \circ \phi) = \frac{1}{\Omega}\sum_{p\in\Omega}{\| I_{t+1}(p)-(I_{t} \circ \phi)(p)) \|^2}, \\
\end{equation}
\begin{equation}
\label{eq:SmoothLoss}
\begin{split}
&L_{smooth}(\phi)=\sum_{p\in\Omega}{||\nabla\phi (p)||^2},
%&\nabla\phi (p) =(\frac{\partial\phi(p)}{\partial x},\frac{\partial\phi(p)}{\partial y},\frac{\partial\phi(p)}{\partial z}),
\end{split}
\end{equation} 
where $p$ is a voxel location (coordinates) in the entire volume, $\Omega$ is a set of the coordinates in the entire volume.
%and $\nabla\phi$ is the spatial gradients of displacement $\phi$.

\subsubsection{Inference}
The purpose of using the estimated displacement map is to warp the detected cell positions but not to align the entire volumes. Therefore, we input the estimated cell position heatmaps instead of the original images, where the details of the cell position heatmap will be described in the next section.
Since the cell position heatmap represents cell positions more clearly compared to the original image, the estimated warped positions are more accurate.

In addition, we perform the test time fine-tuning to obtain accurate warping results since the estimation results of the displacement are not perfect on the test time in general. Since VoxelMorhph \cite{balakrishnan2019voxelmorph} doesn't require supervised data, we can retrain the module in the test phase without using supervised data.
In the test time, given pair images in consecutive frames, we additionally train the network only using the current inputs. It may overfit the training data, but the estimated cell movements become accurate for the consecutive frames. 
%In the test-time fine-tuning, we terminated the training if the $\mathcal{L}_{sim}$ is less than a threshold.
In the test-time fine-tuning, we terminated the training if the magnitude of the updates of the displacement map was less than a threshold.
This fine-tuning is performed for each consecutive frames and obtains the set of the displacement maps $\phi_{t \rightarrow t+1}, (t=1,2,...,K-1)$. The estimated displacement maps will be used for both detection and association.

\subsection{Pairwise Detection for Consistent Detection}
We use the point-based cell detection method \cite{nishimura2019weakly} as the backbone of our cell detection method.
This method estimates a cell position heatmap for a single input image in 2D.
We extend this method for 3D cell detection in the volume and introduce pairwise inputs to achieve consistent detection in consecutive frames.

We first explain the output representation of the cell position heatmap before describing the details of the pairwise inputs.
Given 3D image $I_t$ and the set of annotated cell positions $\{ \bm{c_i} = (x_i, y_i, z_i) \}_{i=1}^{N_t}$ as training data, the ground truth of the cell position heatmap $A_t$ is generated as shown in Fig. \ref{fig:overview}.
In the heatmap, an annotated cell position $\bm{c}_i$ becomes a peak, and the value gradually decreases with a Gaussian distribution. 
The network is trained to produce such a cell position heatmap.
To train the network $d$, we use the mean of the squared error loss function (MSE) defined as:
\begin{equation}
\label{detection_loss}
Loss =\frac{1}{K}\sum_{t=0}^{K} {({A_{t}-\hat{A}_{t})^{2}}},
\end{equation}
where $\hat{A}_{t}$ is the estimation results from the network, and $K$ is the number of the training images.
Then, the peak positions whose intensity is larger than a threshold are detected as cell positions $\{ \hat{\bm{c_i}}^{(t)}\}_{i=1}^{N'_t}$.
For the network, we used U-net \cite{ronneberger2015u}.

To obtain consistent detection results in consecutive frames, we propose a pairwise detection method, which inputs both the original image at $t+1$ and the information of the detection results at $t$.
Since the original position of the detection results at $t$ is misaligned from the cell positions at the next frame $t+1$, cell positions are warped using the displacement map $\phi_{t->t+1}$ estimated by the previous step.
However, if we directly warp the heatmap $\hat{A}_{t}$, the Gaussian distribution gets out of shape and may affect the estimation worse.
We thus re-generate the heatmap $A'_{t \rightarrow t+1}$ based on the warped estimated cell positions, defined as:
\begin{equation}
\hat{\bm{c_i}}^{(t \rightarrow t+1)} =  \hat{\bm{c_i}}^{(t)} + \phi_{t \rightarrow t+1}(\hat{\bm{c_i}}^{(t)}) \hspace{2mm} (i=1, ..., N_{t}),
\end{equation}
where $\phi_{t \rightarrow t+1}(\hat{\bm{c_i}}^{(t)})$ indicates the displacement vector at the position $\hat{\bm{c_i}}^{(t)}$.
Given these two inputs ($A'_{t \rightarrow t+1}$ and $I_{t+1}$), the network $d$ estimates the cell position heatmap $\hat{A}_{t+1}$.
We expect that the network trained using the pairwise inputs produces detection results that contain cell positions corresponding to those at the previous frame, i.e., if a cell is detected at $t$, the corresponding cell is also detected at $t+1$ since the network can know the previous detection results.
The detected cell positions at $t+1$ are represents as $\{ \hat{\bm{c_i}}^{(t+1)}\}_{i=1}^{N'_{t+1}}$.

\subsection{Cell Association using Estimated Cell Motion}
Next, we associate the detection results $\bm{C}^t = \{ \hat{\bm{c_i}}^{(t)}\}_{i=1}^{N'_t}$ at $t$ and $\bm{C}^{t+1} = \{ \hat{\bm{c_i}}^{(t+1)}\}_{i=1}^{N'_{t+1}}$ at $t+1$.
To address large displacement of cell positions, we also use the warped positions $\bm{C}^{(t \rightarrow t+1)} = \{ \hat{\bm{c_i}}^{(t \rightarrow t+1)}\}_{i=1}^{N'_t}$.

Given these two sets of cell positions $\bm{C}^{(t \rightarrow t+1)}$ and $\bm{C}^{t+1}$ in consecutive frames, we solve the one-by-one matching problem using linear programming.
A set of all association hypotheses between the cell region is listed up, where if the distance between $\hat{\bm{c_i}}^{(t \rightarrow t+1)}$ and $\hat{\bm{c_j}}^{(t+1)}$ is less than a threshold, the association $i \rightarrow j$ is added to the hypothesis.
The optimization for selecting the optimal set of association hypotheses can be solved by binary programming that has constraints to avoid conflict association. The optimization is formalized as:
\begin{equation}
\bm{x^*} = \arg \max_{\bm{x}} {\bm{\rho}^T \bm{x}} \hspace{3mm} s.t. \hspace{2mm} G^T \bm{x} \leq \bm{1}, \hspace{2mm} \bm{x} \in \{0,1\},
\end{equation}
where vector $\bm{\rho}$ stores the score of every hypothesis and matrix $G$ stores the constraints to avoid conflict hypothesis, which is defined as similar to \cite{bise2011reliable}. This problem can be relaxed to linear programming.

%\begin{equation}
%\label{farame_by_frame}
%P_{link}(b_j|c_i) =\exp{(\frac{||f(c_i)-f(b_j)||}{σ})}　
%\end{equation}
 
\section{Experiments}
\subsection{Datasets and Implementation Details}
In the experiments, we used four time-lapse sequences captured by a confocal microscope, where neuron cells of C. elegans move during scanning. The frame rates in Seq. 1, 2, and 3 are 1.5 frames per secs (fps) that in Seq. 4 is 2 fps, and the number of frames is 150 in all sequences with $128\times576\times32$ pixels image resolution.

%Each sequence has a different cell motion, which depends on the characteristics of the individuals; In Seq. 1 and 4, the cell motion is very large; In Seq. 3 and 4, the cell motion is not so large. 
The ground truth of cell trajectories is generated by manual tracking using Ilastik \cite{berg2019ilastik}. The cell trajectories are represented as (CellID, frame, x, y, z).
The average number of individual cells in each sequence is about 120, and the total number of annotated cells is 72,716 in all sequences. 
We evaluated the proposed method using leave-one-out (four-fold cross-validation). We used Tracking Accuracy (TA) and Target Effectiveness (TE) \cite{kanadeT2011} as tracking performance metrics. The tracking accuracy is the number of true positive associations divided by that in the ground truth, and the target effectiveness is the number of associations that continuously succeed over the total number of frames of the target.

We implemented our method by using PyTorch.
%\cite{paszke2019pytorch}.
To train the detection and alignment networks, we used the ADAM optimizer %\cite{kingma2014adam}
with a learning rate of $10^{-3}$, $10^{-2}$, epoch $=20, 1500$, mini-batch size $=4,8$, respectively. The thresholds for cell detection and association were 0.05 and 10 pixels, respectively, and $\gamma=0.01$.

\begin{table}[t]
    \def\@captype{table}
      \makeatother
        \centering
        \caption{Tracking performance in terms of Tracking Accuracy (TA) and Target Effectiveness (TE) of comparative methods.} 
        \begin{tabular}{c|ccc|cc} 
        \hline
        Method & R & FT & PD & TA & TE \\ \hline\hline
        w/o R, FT, PD  & & & & 0.7927 & 0.6522 \\
        w/o FT, PD & $\checkmark$ & & & 0.7975 & 0.6630 \\
        w/o PD & $\checkmark$ & $\checkmark$ & & 0.8088 & 0.7361 \\
        Ours   & $\checkmark$ & $\checkmark$ & $\checkmark$ & \textbf{0.8302} & \textbf{0.7466} \\
        \hline
        \end{tabular}
        \label{tab:trackingAcc}
\end{table}

\begin{figure}[t]
 \begin{center} \includegraphics[width=0.9\columnwidth]{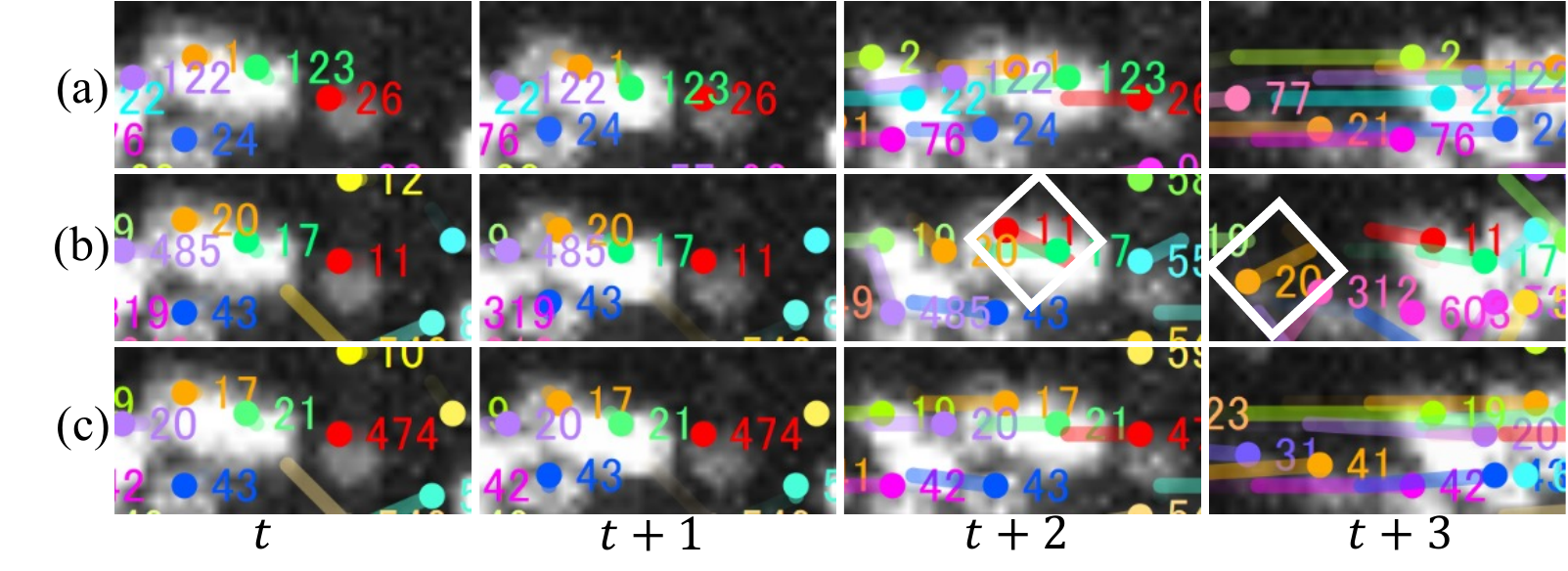}
    \caption{Example of tracking results of (a) Ground truth, (b)w/o registration, and (c) our method. The same cell trajectory has the same color and ID. `$\diamond$' indicates a switching error.
    %Tracking results from each compared methods.(a)Ground truth,(b)w/o redistration,(c)w/ registration\\(a)ground truth (b)w/o registration (c)test time registration
    }
    \label{fig:tracking_result}
 \end{center}
 \vspace{-5mm}
\end{figure}

\subsection{Quantitative Evaluation of Tracking Performance}
To show the effectiveness of each module of the proposed method, we compared our method with the three methods: 1) (w/o R, FT, PD), which is a baseline method that first detects individual cells by V-net only using a single image input, associates the detection results by linear programming, where `R', `FT', `PD' indicate `registration', `fine-tuning', and `pairwise detection', respectively. This baseline is similar to \cite{lauziere2022semi}; 2) (w/o FT, PD), which only used registration without fine-tuning; 3) (w/o PD), which did not use pairwise detection.

Table \ref{tab:trackingAcc} shows the tracking performances (TA, TE) of comparative methods.
In these results, every module improved the tracking performances incrementally, and the proposed method achieved the best performance in this comparison.

Fig. \ref{fig:tracking_result} shows the example of tracking results. In (b) the baseline method, switching errors often occurred since association cannot work properly due to large displacement. In contrast, our method properly associated corresponding cells by the effect of cell position heatmap alignment.

\begin{figure}[t]
 \begin{center}
\includegraphics[width=0.92\columnwidth]{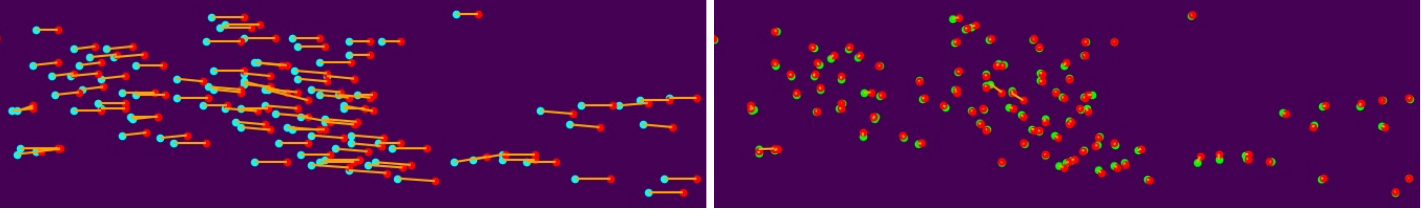}
    \caption{Examples of warped positions by cell position heatmap-based alignment. Left: original displacement. Right: displacement after warping. blue is the position at $t$, green is the warped position from $t$ to $t+1$, red is that at $t+1$, and the line is the displacement.}
    \label{Fig:registration}
 \end{center}
 \vspace{-3mm}
\end{figure}

Fig. \ref{Fig:registration} shows an example of motion estimation by cell position heatmap alignment. Before warping, the displacement is large; however, the distances between corresponding cells are closed after alignment. This facilitates cell association and improves tracking performance.

 \begin{figure}[t]
 \begin{center}
\includegraphics[width=0.80\columnwidth]{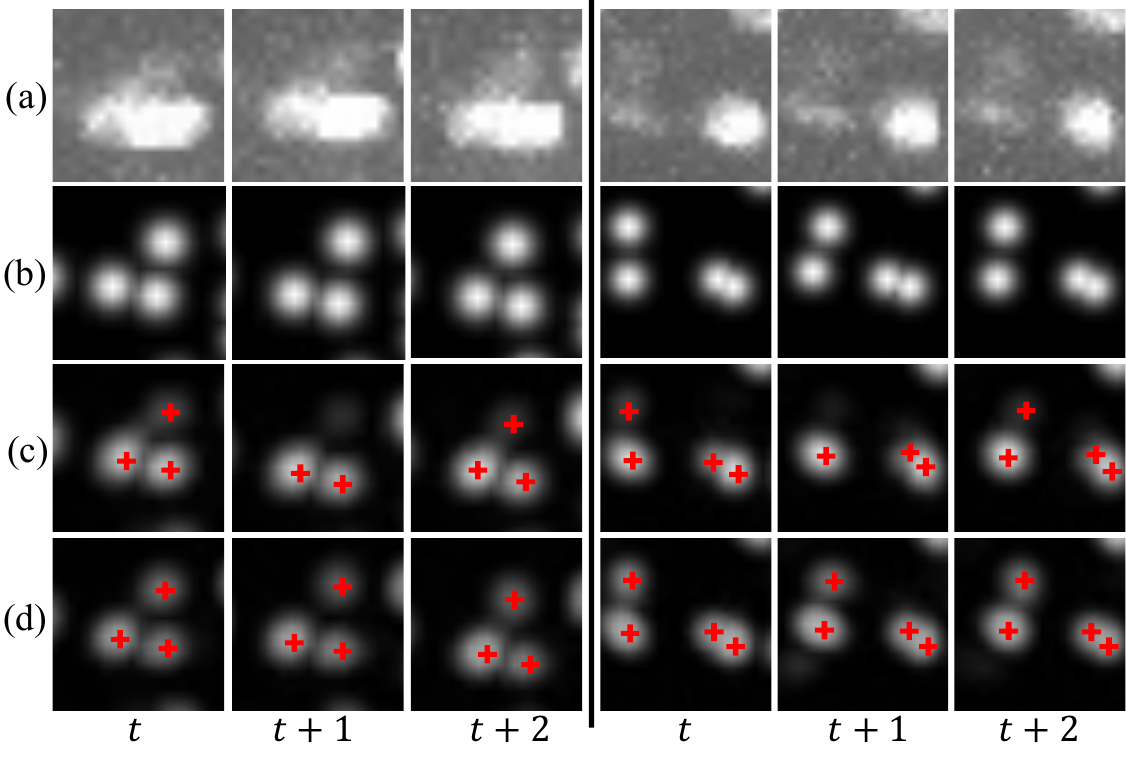}
    \caption{Examples of effectiveness of pairwise detection. 
    (a)Original image, (b)Ground truth, (c)Single detection, and (d)Pairwise detection.}
    \label{fig:detection}
 \end{center}
  \vspace{-5mm}
\end{figure}

% pairwise detection
Fig. \ref{fig:detection} shows two examples of the detection results in consecutive frames.
In the left example, there are three cells, where two cells touched and made a cluster, and one cell becomes very low contrast. In the right example, there are four cells, where two cells are very low contrast, and two cells are touched.
In (c), estimated by a baseline detection method that inputs a single image, the low contrast cell could not be detected at $t+1$ in both cases, i.e., the detection is inconsistent in consecutive frames. This inconsistent detection may cause tracking errors.
In contrast, our method detects all cells consistency. We consider this consistent detection contributes to improving the tracking performance.

\section{CONCLUSIONS}
We proposed a cell tracking method in {\it C. elegans}, which addresses two main issues: large displacement and inconsistent detection in consecutive frames.
First, we introduce cell position heatmap-based alignment to estimate the cell positions at the next frame, which facilitates accurate cell association.
Second, we propose a pairwise detection, which can use the detection results at the previous frame for detection at the current frame. This produces more consistent detection results.
The experimental results demonstrated the effectiveness of each module, and the proposed method achieved the best performance in comparison.

\addtolength{\textheight}{-12cm}   % This command serves to balance the column lengths

%%%%%%%%%%%%%%%%%%%%%%%%%%%%%%%%%%%%%%%%%%%%%%%%%%%%%%%%%%%%%%%%%%%%%%%%%%%%%%%%

%%%%%%%%%%%%%%%%%%%%%%%%%%%%%%%%%%%%%%%%%%%%%%%%%%%%%%%%%%%%%%%%%%%%%%%%%%%%%%%%

%%%%%%%%%%%%%%%%%%%%%%%%%%%%%%%%%%%%%%%%%%%%%%%%%%%%%%%%%%%%%%%%%%%%%%%%%%%%%%%%
%\section*{ACKNOWLEDGMENT}
\noindent {\bf Acknowledgments:}
This work was supported by JSPS KAKENHI Grant Number JP20H04211, Japan and AMED under Grant Number JP19he2302002

%%%%%%%%%%%%%%%%%%%%%%%%%%%%%%%%%%%%%%%%%%%%%%%%%%%%%%%%%%%%%%%%%%%%%%%%%%%%%%%%

%References are important to the reader; therefore, each citation must be complete and correct. If at all possible, references should be commonly available publications.

\bibliographystyle{IEEEbib}
\bibliography{refer}
%\begin{thebibliography}{99}

%\bibitem{c1} G. O. Young, ÒSynthetic structure of industrial plastics (Book style with paper title and editor),Ó 	in Plastics, 2nd ed. vol. 3, J. Peters, Ed.  New York: McGraw-Hill, 1964, pp. 15Ð64.

%\bibitem{c2} Balakrishnan, Guha and Zhao, Amy and \textit{et al.},"An Unsupervised Learning Model for Deformable Medical Image Registration,".in CVPR 2018

%\end{thebibliography}

\end{document}